\documentclass[runningheads]{llncs}
\usepackage{tikz}
\usepackage{graphicx}
\usepackage{soul}
\usepackage{url}
\usepackage[hidelinks]{hyperref}
\usepackage[small]{caption} 
\usepackage{graphicx}
\usepackage{amsmath}
\usepackage{booktabs}
\usepackage{listings}
\usepackage{amsmath,amssymb}
\usepackage[ruled,vlined,linesnumbered,nokwfunc]{algorithm2e}
\usepackage{color}
\usepackage[bbgreekl]{mathbbol}
\usepackage[mathscr]{euscript}

\newcommand{\code}[1]{\texttt{#1}}
\newcommand{\eofex}{\mbox{}\nobreak\hfill\hspace{0.5em}$\blacksquare$}
\newcommand{\system}[1]{\textsc{#1}}

\newcommand{\rg}[3]{#1{#2\,}\ldots{#2\,}#3}
\newcommand{\set}[1]{\{#1\}}
\newcommand{\pair}[2]{\langle{#1},{#2}\rangle}
\newcommand{\true}{\top}

\def\IA{\mathsf{IA}}

\def\IA{\mathsf{IA}}

\DeclareFontFamily{U} {MnSymbolC}{}
\DeclareFontShape{U}{MnSymbolC}{m}{n}{
  <-6> MnSymbolC5
  <6-7> MnSymbolC6
  <7-8> MnSymbolC7
  <8-9> MnSymbolC8
  <9-10> MnSymbolC9
  <10-12> MnSymbolC10
  <12-> MnSymbolC12}{}
\DeclareFontShape{U}{MnSymbolC}{b}{n}{
  <-6> MnSymbolC-Bold5
  <6-7> MnSymbolC-Bold6
  <7-8> MnSymbolC-Bold7
  <8-9> MnSymbolC-Bold8
  <9-10> MnSymbolC-Bold9
  <10-12> MnSymbolC-Bold10
  <12-> MnSymbolC-Bold12}{}

\DeclareSymbolFont{MnSyC} {U} {MnSymbolC}{m}{n}

\DeclareMathSymbol{\blackdiamond}{\mathbin}{MnSyC}{109}

\def\QCN{\mathsf{QCN}}
\def\D{{\mathsf{D}}}
\def\B{{\mathsf{B}}}

\def\IA{{\mathsf{IA}}}

\usepackage{graphicx}
\usepackage{url}
\usepackage{footmisc}
\usepackage{paralist}
\usepackage{array}
\usepackage{subcaption}
\newcolumntype{L}[1]{>{\raggedright\let\newline\\\arraybackslash\hspace{0pt}}m{#1}}
\newcolumntype{C}[1]{>{\centering\arraybackslash}m{#1}}
\newcolumntype{N}{@{}m{0pt}@{}}
\newcolumntype{R}[1]{>{\raggedleft\let\newline\\\arraybackslash\hspace{0pt}}m{#1}}

\SetKwInOut{In}{in}
\SetKwInOut{Out}{out}
\SetKwInOut{InOut}{in/out}
\SetKwInOut{Input}{input}
\SetKwInOut{Output}{output}
\SetKw{Break}{break}
\SetKwRepeat{Do}{do}{while}

\begin{document}
\title{Allen's Interval Algebra Makes the Difference}
\author{Tomi Janhunen\inst{1,2}\orcidID{0000-0002-2029-7708} \and
Michael Sioutis\inst{1}\orcidID{0000-0001-7562-2443}}
\authorrunning{T. Janhunen and M. Sioutis}
\institute{Department of Computer Science, Aalto University, Espoo, Finland \and Computing Sciences Unit, Tampere University, Tampere, Finland\\
\email{firstname.lastname@aalto.fi}}

\maketitle

\begin{abstract}
Allen's Interval Algebra constitutes a framework for reasoning about
temporal information in a qualitative manner. In particular, it uses
intervals, i.e., pairs of endpoints, on the timeline to represent
entities corresponding to actions, events, or tasks, and binary
relations such as \emph{precedes} and \emph{overlaps} to encode the
possible configurations between those entities. Allen's calculus has
found its way in many academic and industrial applications that
involve, most commonly, planning and scheduling, temporal databases,
and healthcare.
In this paper, we present a novel encoding of Interval Algebra
using answer-set programming (ASP) extended by difference constraints,
i.e., the fragment abbreviated as ASP(DL), and demonstrate its
performance via a preliminary experimental evaluation.  Although our
ASP encoding is presented in the case of Allen's calculus for the sake
of clarity, we suggest that analogous encodings can be devised for
other point-based calculi, too.
\keywords{%
Answer Set Programming,
Difference Constraints,
Qualitative Constraints,
Spatial and Temporal Reasoning,
Symbolic AI}
\end{abstract}


\section{Introduction}

Qualitative Spatial and Temporal Reasoning (QSTR) is a Symbolic AI approach
that deals with the fundamental cognitive concepts of space and time in a qualitative, human-like, manner~\cite{DBLP:journals/csur/DyllaLM0DVW17,ligozat2013qualitative}.
As an illustration, the first constraint language to deal with time on a qualitative level was proposed
by Allen in~\cite{DBLP:journals/cacm/Allen83}, called Interval Algebra.
Allen wanted to define
a framework for reasoning about time in the context of natural language processing that would be reliable and efficient enough for reasoning
about temporal information in a qualitative manner. In particular, Interval Algebra uses intervals on the timeline to represent
entities corresponding to actions, events, or tasks, and relations such as \emph{precedes} and \emph{overlaps} to encode the possible configurations between those entities. 
Interval Algebra has become one of the most well-known
qualitative constraint languages, due to its use for representing and reasoning about temporal information in various applications.
More specifically, typical applications of Interval Algebra involve
planning and scheduling~%
\cite{DBLP:conf/kr/Allen91,DBLP:conf/ijcai/AllenK83,DBLP:journals/dke/Dorn95,DBLP:conf/icra/MudrovaH15,DBLP:conf/aaai/PelavinA87},
natural language processing~%
\cite{DBLP:conf/ijcai/DenisM11,DBLP:conf/aaai/SongC88},
temporal databases~%
\cite{DBLP:conf/deductive/ChenZ98,DBLP:journals/tods/Snodgrass87},
multimedia databases~\cite{DBLP:journals/tkde/LittleG93},
molecular biology~\cite{DBLP:journals/jacm/GolumbicS93}
(e.g., arrangement of DNA segments/intervals along a linear chain
involves particular temporal-like problems~\cite{benzer59}),
workflow~\cite{DBLP:conf/adc/LuSPG06}, and
healthcare~%
\cite{DBLP:journals/datamine/KostakisP17,DBLP:journals/kais/MoskovitchS15a,qr2017}. 

Answer-set programming (ASP) is a declarative programming paradigm
\cite{BET11:jacm,JN16:aimag} designed for solving computationally hard
search and optimization problems from the first two levels of
polynomial hierarchy.  Typically, one \emph{encodes} the solutions of
a given problem as a logic program and then uses an answer-set solver
for their computation. The idea of representing Allen's Interval
Algebra in terms of rules is not new; existing encodings can be found
in \cite{BFB16:iclp,Li12:ictai}.
However, these encodings do not scale well when the number of
intervals is increased beyond $20$~\cite[Section~$6$]{BFB16:iclp}. The
likely culprit for decreasing performance is the explicit
representation of compositions of base relations, which tends to cause
cubic blow-ups when instantiating the encoding for a particular
problem instance.
In this paper, we circumvent such negative effects by using an appropriate
extension of ASP to encode the underlying constraints of Allen's
calculus.  The crucial primitive is provided by difference logic (DL)
\cite{NO05:cav} featuring \emph{difference constraints} of form
$x-y\leq k$. The respective fragment of ASP is known as ASP(DL)
\cite{JKOSWS17:tplp} and it has been efficiently implemented within
the \system{clingo} solver family. When encoding Allen's calculus in
ASP(DL), the transitive effects of relation composition can be
delegated to propagators implementing difference constraints.  Hence,
no blow-ups result when instantiating the ASP rules for a particular
constraint network and the resulting ground logic program remains
linear in network size.

The rest of this article is organized as follows. The basic notions of
qualitative constraint networks ($\QCN$s) and, in particular, Allen's
Interval Algebra are first recalled in
Section~\ref{sec:preliminaries}. Then, difference constrains are
introduced in Section~\ref{sec:asp-dl} and we also show how they are
available in ASP, i.e., the fragment abbreviated as ASP(DL).  The
actual encodings of $\QCN$s in ASP(DL) are presented in
Section~\ref{sec:encodings}. The preliminary experimental evaluation
of the resulting encodings takes place in
Section~\ref{sec:experiments}.  Finally, we present our conclusions
and future directions in Section~\ref{sec:conclusion}.


\section{Preliminaries}
\label{sec:preliminaries}

A binary qualitative constraint language is based on a finite set 
$\B$ of \emph{jointly exhaustive and pairwise disjoint} relations, called the set of \emph{base relations}~\cite{DBLP:conf/pricai/LigozatR04}, that is defined over an infinite domain $\D$. 
These base relations represent definite
knowledge between two entities with respect to the level of granularity provided by the domain $\D$; 
indefinite knowledge can
be specified by a union of possible base relations, and is represented by the set containing them. The set $\B$ contains the identity relation ${\mathsf{Id}}$,
and is closed under the \emph{converse} operation ($^{-1}$). The total set of relations $2^{\B}$ is equipped with the usual set-theoretic operations of union and intersection,
the converse operation, and the \emph{weak composition} operation denoted by $\diamond$~\cite{DBLP:conf/pricai/LigozatR04}. 
For all $r \in 2^\B$, $r^{-1} = \bigcup\{b^{-1}~|~b \in r\}$. 
The weak composition~$(\diamond)$ of two base relations $b,b^\prime \in \B$
is defined as the smallest (i.e., strongest) relation $r \in 2^{\B}$ that includes $b \circ b^\prime$, or, formally,
$b \diamond b^\prime {=} \{ b^{\prime\prime} \in \B~|~b^{\prime\prime}{\cap}(b \circ b^\prime) \neq \emptyset \}$, where
$b \circ b^\prime {=} \{ (x, y) \in \D\times{\D}~|~\exists{z \in \D}~\text{such that}~( x, z) \in b \wedge (z, y) \in b^\prime \}$ is the (true) composition of $b$ and $b^\prime$.
For all $r,r^\prime \in 2^\B$, $r \diamond r^\prime$ $=$  $\bigcup\{b\diamond{b^\prime}~|~b\in{r},b^\prime\in{r^\prime}\}$. 

As an illustration, consider the well-known qualitative temporal constraint language of Interval Algebra ($\IA$),
introduced by Allen in~\cite{DBLP:journals/cacm/Allen83}.
The domain $\D$ of Interval Algebra is defined to be the set of intervals on the line of rational numbers,
i.e., $\D$ $=$ $\{x=(x^-, x^+) \in \mathbb{Q}\times\mathbb{Q}~|~x^-<x^+\}$.
Each base relation can be defined by appropriately constraining the endpoints of the two intervals at hand, which yields a total of $13$ base relations comprising
the set $\B=\{e$, $p$, $pi$, $m$, $mi$, $o$, $oi$, $s$, $si$, $d$, $di$, $f$, $fi\}$; 
these symbols are explained in the caption of Figure~\ref{figA}. For example, $d$
is defined as $d$ $=$ $\{(x,y)\in{\D\times{\D}}~|~x^->y^-~\text{and}~x^+<y^+\}$. The identity relation 
$\mathsf{Id}$ of Interval Algebra is $e$ and its converse is again $e$. 

\begin{figure}[t!]
\centering
\hspace*{\fill}%
\begin{subfigure}[b]{0.45\textwidth}
\centering
        \begin{tikzpicture}[scale=.6]
          \node[draw,circle] (x1) at (0,0) {$x_1$};
          \node[draw,circle] (x2) at (4,0) {$x_2$};
          \node[draw,circle] (x3) at (4,4) {$x_3$};
          \node[draw,circle] (x4) at (0,4) {$x_4$};
          \draw [->] (x1) -- (x2) node [midway, fill=white] {$\{p,m\}$};
          \draw [dotted,-] (x1) -- (x3) node [near end, fill=white] {$\B$};
          \draw [->] (x1) -- (x4) node [midway, fill=white] {$\{d,s,fi\}$};
          \draw [->] (x2) -- (x3) node [midway, fill=white] {$\{oi\}$};
          \draw [->] (x2) -- (x4) node [near start, fill=white] {$\{oi,m\}$};
          \draw [->] (x3) -- (x4) node [midway, fill=white] {$\{pi,e\}$};
          
         \end{tikzpicture}
\caption{A satisfiable $\QCN$ ${\mathcal N}$\label{figA:B}}
\end{subfigure}
\hfill%
\begin{subfigure}[b]{0.45\textwidth}
\centering
        \begin{tikzpicture}[scale=.6]
          \draw [->] (0,0) -- (5,0);
          \draw  node[fill,circle,inner sep=0pt,minimum size=3pt] (x1) at (0.5,0) {};
          \draw  node[fill,circle,inner sep=0pt,minimum size=3pt] (x2) at (1.25,0) {};
          \draw  node[fill,circle,inner sep=0pt,minimum size=3pt] (x3) at (2.5,0) {};
          \draw  node[fill,circle,inner sep=0pt,minimum size=3pt] (x4) at (3.25,0) {};
          \draw  node[fill,circle,inner sep=0pt,minimum size=3pt] (x5) at (4.25,0) {};

          \draw  node[fill, circle,inner sep=0pt,minimum size=1pt] (v1) at (1.25,4) {};
          \draw  node[fill, circle,inner sep=0pt,minimum size=1pt] (v2) at (2.5,4) {};
          \draw [-] (v1) -- (v2) node [midway,above] {$x_1$};

          \draw  node[fill, circle,inner sep=0pt,minimum size=1pt] (v3) at (2.5,3) {};
          \draw  node[fill, circle,inner sep=0pt,minimum size=1pt] (v4) at (4.25,3) {};
          \draw [-] (v3) -- (v4) node [midway,above] {$x_2$};

          \draw  node[fill, circle,inner sep=0pt,minimum size=1pt] (v5) at (0.5,2) {};
          \draw  node[fill, circle,inner sep=0pt,minimum size=1pt] (v6) at (3.25,2) {};
          \draw [-] (v5) -- (v6) node [midway,above] {$x_3$}; 

          \draw  node[fill, circle,inner sep=0pt,minimum size=1pt] (v7) at (0.5,1) {};
          \draw  node[fill, circle,inner sep=0pt,minimum size=1pt] (v8) at (3.25,1) {};
          \draw [-] (v7) -- (v8) node [midway,above] {$x_4$}; 

          \draw [dotted,-] (v1) -- (x2);
          \draw [dotted,-] (v2) -- (x3);
          \draw [dotted,-] (v3) -- (x3);
          \draw [dotted,-] (v4) -- (x5);
          \draw [dotted,-] (v5) -- (x1);
          \draw [dotted,-] (v6) -- (x4);
          \draw [dotted,-] (v7) -- (x1);
          \draw [dotted,-] (v8) -- (x4);

         \end{tikzpicture}
\caption{A solution $\sigma$ of ${\mathcal N}$\label{solorig}}
\end{subfigure}
\hspace*{\fill}
\caption{Examples of $\QCN$ terminology using Interval Algebra; symbols $p$, $e$, $m$, $o$, $d$, $s$, and $f$ correspond to
the base relations \emph{precedes}, \emph{equals}, \emph{meets}, \emph{overlaps}, \emph{during}, \emph{starts}, and \emph{finishes} respectively, 
with $\boldsymbol{\cdot}{i}$ denoting the converse of $\boldsymbol{\cdot}$ (note that $ei$ $=$ $e$)}
\label{figA}
\end{figure}
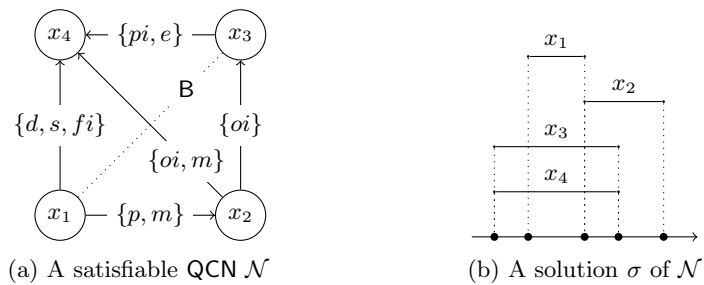
\begin{definition}\label{def:qcn}A \emph{qualitative constraint network ($\mathsf{QCN}$)}
is a tuple $(V,C)$ where:
\begin{compactitem}
\item $V$ $=$ $\{v_1,$ $\ldots,$ $v_n\}$ is a non-empty finite set of variables, each representing an entity of an infinite domain $\D$; 

\item and $C$ is a mapping $C:V\times{V}\rightarrow{2^\B}$ such that $C(v,v) = \{\mathsf{Id}\}$ for all $v \in V$ and $C(v,v^\prime)=(C(v^\prime,v))^{-1}$ for all $v,v^\prime \in V$.

\end{compactitem}
\end{definition}

An example of a $\QCN$ of $\IA$ is shown in Figure~\ref{figA:B}; for
clarity, neither converse relations nor $\mathsf{Id}$ loops are
mentioned or shown in the figure.

Given a $\QCN$ ${\mathcal N} = (V,C)$, a \emph{solution} of
${\mathcal N}$ is a mapping $\sigma:V\rightarrow\D$ such that
$\forall(u,v) \in V\times{V}$, $\exists{b} \in C(u,v)$ so that
$(\sigma(u),\sigma(v)) \in b$ (see Figure~\ref{solorig}).


\section{Difference Constraints for Answer-Set Programming}
\label{sec:asp-dl}

We assume that the reader is already familiar with the basics of ASP
(cf. \cite{BET11:jacm,JN16:aimag}) and merely concentrate on extending
ASP in terms of \emph{difference constraints}. Such a constraint is an
expression of the form $x-y\leq k$ where $x$ and $y$ are variables and
$k$ is a constant. Intuitively, the \emph{difference} of $x$ and $y$
should be less than or equal to $k$. Potential domains for $x$ and $y$
are integers and reals, for instance. The domain is usually determined
by the application and, for the purposes of this paper, the set of
integers is assumed in the sequel.
The given form of difference constraints can be taken as a normal form
for such constraints. However, with a little bit of elaboration some
other and very natural constraints concerning $x$ and $y$ become
expressible. While $x\leq y$ is equivalent to $x-y\leq 0$, the strict
difference $x<y$ translates into $x-y\leq-1$. To state the equality
$x=y$, two difference constraints emerge, since $x=y$ $\iff$
$x-y\leq 0$ and $y-x\leq 0$.

Difference constraints can be implemented very efficiently, since they
enable a linear-time check for unsatisfiability. Given a set $S$ of
such constraints, one can use the Bellman-Ford algorithm to check if
$S$ has a \emph{loop} of variables $\rg{x_1}{,}{x_n}$ where $x_n=x_1$
along with difference constraints
$\rg{x_2-x_1\leq d_1}{,}{x_n-x_{n-1}\leq d_{n-1}}$
such that $\sum_{i=1}^{n-1}d_i<0$. When carrying out the check
for satisfiability, it is not necessary to find concrete values for the
variables in $S$. This is in perfect line with the idea of reasoning
about $\QCN$s on a qualitative, symbolic, level.

\begin{example}
The set of difference constraints
$S_1=\set{y-x\leq 1,z-y\leq 1,x-z\leq -3}$
is unsatisfiable, since $1+1-3<0$. However, if the second difference
constraint is revised to $z-y\leq 2$, the resulting set of difference
constraints $S_2$ is satisfiable, as witnessed by an assignment with
$x=0$, $y=1$, and $z=3$.
\eofex
\end{example}

More formally, an \emph{assignment} $\tau$ is a mapping from variables
to integers and a difference constraint $x-y\leq k$ is
\emph{satisfied} by $\tau$, denoted $\tau\models x-y\leq k$, if
$\tau(x)-\tau(y)\leq k$. Also, we write $\tau\models S$ for a set of
difference constraints $S$, if $\tau\models x-y\leq k$ for every
constraint $x-y\leq k$ in $S$. If $\tau\models S$, we also say that
$S$ is \emph{satisfiable} and that $\tau$ is a \emph{solution} to $S$.
Moreover, it is worth pointing out that if $\tau\models S$ then also
$\tau'\models S$ where $\tau'(x)=\tau(x)+k$ for some integer $k$. Thus
$S$ has infinitely many solutions if it has at least one solution. If
$S$ is satisfiable, it is easy to compute one concrete solution by
using a particular variable $z$ as a point of reference via the
intuitive assignment $\tau(z)=0$.
\footnote{This distinguished variable $z$ can be used as a name for $0$ in
  other difference constraints. Then, e.g., $x-z\leq k$ and $z-x\leq -k$
  express together that $x=k$.}

Difference logic (DL) extends classical propositional logic in the
\emph{satisfiability modulo theories} (SMT) framework \cite{NO05:cav}.
A propositional formula $\phi$ in DL is formed in terms of usual
atomic propositions $a$ and difference constraints $x-y\leq k$.  A
\emph{model} of $\phi$ is a pair $\pair{\nu}{\tau}$ such that
(i)
$\nu,\tau\models a$ iff $\nu(a)=\true$,
(ii)
$\nu,\tau\models x-y\leq k$ iff $\tau\models x-y\leq k$, and
(iii)
$\nu,\tau\models\phi$ by the recursive rules of propositional logic.
Difference logic lends itself for applications where integer variables
are needed in addition to Boolean ones. Thus, it serves as a potential
target formalism when it comes to implementing ASP via translations
\cite{Janhunen18:ki,JNS09:lpnmr}.

The rule-based language of ASP can be generalized in an analogous way
by using difference constraints as additional conditions in rules. The
required theory extension of the \system{clingo} solver is
documented in \cite{GKKOSW16:iclp}. For instance, a difference
constraint $x-y\leq 5$ can be expressed as
\code{\&diff\{x-y\} <= 5}
where \code{x} and \code{y} are constants in the syntax of ASP but
understood as integer variables of difference logic. However, using
such fixed names for variables is often too restrictive from
application perspective. It is possible to use function symbols to
introduce collections of integer variables for a particular
application. For instance, if the arcs of a digraph are represented by
the predicate \code{arc/2}, we could introduce a variable
\code{w(X,Y)} for the \emph{weight} for each pair of first-order
variables \code{X} and \code{Y} satisfying \code{arc(X,Y)}.  Recall
that free variables in rules are universally quantified in ASP. More
details about the theory extension corresponding to difference logic
can be found in \cite{JKOSWS17:tplp} whereas its implementation is
known as the \system{clingo-dl} solver.%
\footnote{\url{https://potassco.org/labs/clingodl/}}


\section{Encoding Temporal Networks in ASP(DL)}
\label{sec:encodings}

In what follows, we present our novel encoding of temporal networks
using ASP extended by difference constraints. To encode base relations
from $\B$ in a systematic fashion, we introduce constants
\code{eq},
\code{p},
\code{pi},
\code{m},
\code{mi},
\code{o},
\code{oi},
\code{s},
\code{si},
\code{d},
\code{di},
\code{f}, and
\code{fi}
as names for the base relations (see again Section~\ref{sec:preliminaries}).
The structure of networks themselves is described in terms of
predicate \code{brel/3} whose first two arguments are variables from
the network and the third argument is one possible base relation for
the pair of variables in question. Then, for instance, the base
relations associated with variables $x_1$ and $x_2$ in
Figure~\ref{figA:B} could be encoded in terms of facts
\code{brel(1,2,p)} and \code{brel(1,2,m)}.
Given any such collection of facts, some basic inferences are made
using the ASP rules in Listing~\ref{code:domains}. First, the rules in
lines \ref{line:var1}--\ref{line:var2} extract the identities of
variables for later reference. Secondly, the rule in line \ref{line:arc}
defines the arc relation for the underlying digraph of the network.
Given these pieces of information, we are ready to
formalize the solutions of the temporal network.
For each interval \code{X}, we introduce integer variables
\code{sp(X)} and \code{ep(X)} to capture the respective
\emph{starting} and \emph{ending} points of the interval. The relative
order of theses points is then determined using the difference constraint
expressed by the rule in line \ref{line:sp-ep}. Interestingly, there
is no need to constrain the domain of time points otherwise, e.g.,
by specifying lower and upper bounds; arbitrary integer values are
assumed. In addition, the choice rule in line \ref{line:choose-base}
picks exactly one base relation for each arc of the constraint network.

\begin{lstlisting}[label=code:domains,float=t,%
caption={Choice of Base Relations},escapechar=|]
% Domains
var(X) :- brel(X,Y,R). |\label{line:var1}|
var(Y) :- brel(X,Y,R). |\label{line:var2}|
arc(X,Y) :- brel(X,Y,R). |\label{line:arc}|

% Intervals for every variable X: sp(X) <= ep(X)
&diff{ sp(X)-ep(X) } <= 0 :- var(X). |\label{line:sp-ep}|

% Choose base relations
{ chosen(X,Y,R): brel(X,Y,R) } = 1 :- arc(X,Y). |\label{line:choose-base}|
\end{lstlisting}

\begin{lstlisting}[label=code:relations,float=t!,%
caption={Difference Constraints Expressing Base Relations},escapechar=|]
% Relation eq(X,Y): sp(X) = sp(Y) and ep(X) = ep(Y)
&diff{ sp(X)-sp(Y) } <= 0 :- chosen(X,Y,eq). |\label{line:eq-sp1}|
&diff{ sp(Y)-sp(X) } <= 0 :- chosen(X,Y,eq). |\label{line:eq-sp2}|
&diff{ ep(X)-ep(Y) } <= 0 :- chosen(X,Y,eq). |\label{line:eq-ep1}|
&diff{ ep(Y)-ep(X) } <= 0 :- chosen(X,Y,eq). |\label{line:eq-ep2}|
  
% Relation during(X,Y): sp(Y) < sp(X) and ep(X) < ep(Y)
&diff{ sp(Y)-sp(X) } <= -1 :- chosen(X,Y,d). |\label{line:during-sp}|
&diff{ ep(X)-ep(Y) } <= -1 :- chosen(X,Y,d). |\label{line:during-ep}|
\end{lstlisting}

The satisfaction of the chosen base relations is enforced by further
difference constraints, which are going to be detailed next.  Rather
than covering all $13$, we picked two representatives for more
detailed discussion (see Listing~\ref{code:relations}). In case of
equality, the starting and ending points of intervals \code{X} and
\code{Y} must coincide.  The difference constraints introduced in
lines \ref{line:eq-sp1}--\ref{line:eq-sp2}, whenever activated by the
satisfaction of \code{chosen(X,Y,eq)}, enforce the equality of the
starting points and those of lines
\ref{line:eq-ep1}--\ref{line:eq-ep2} cover the respective ending
points.
The case of the \emph{during} relation is simpler since the
relationships of starting/ending points are strict and only two rules
are needed for a pair of intervals \code{X} and \code{Y}. The rule in
line~\ref{line:during-sp} orders the starting points. The rule in
line~\ref{line:during-ep} puts the ending points in the opposite
order. The encodings for the remaining base relations are obtained
similarly.


\section{Experimental Evaluation}
\label{sec:experiments}

\begin{figure}[t!]
\begin{center}
\begin{tabular}{@{}c@{}c@{}}
\includegraphics[width=0.5\textwidth]{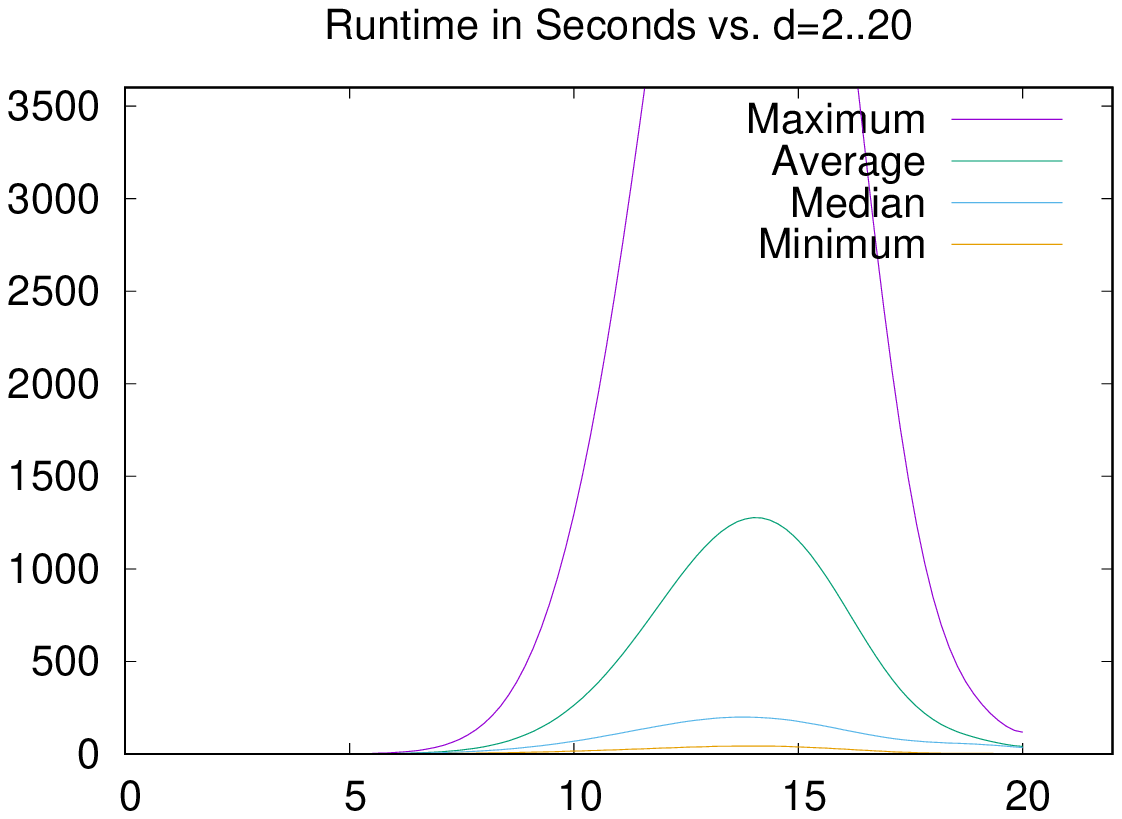}
&
\includegraphics[width=0.5\textwidth]{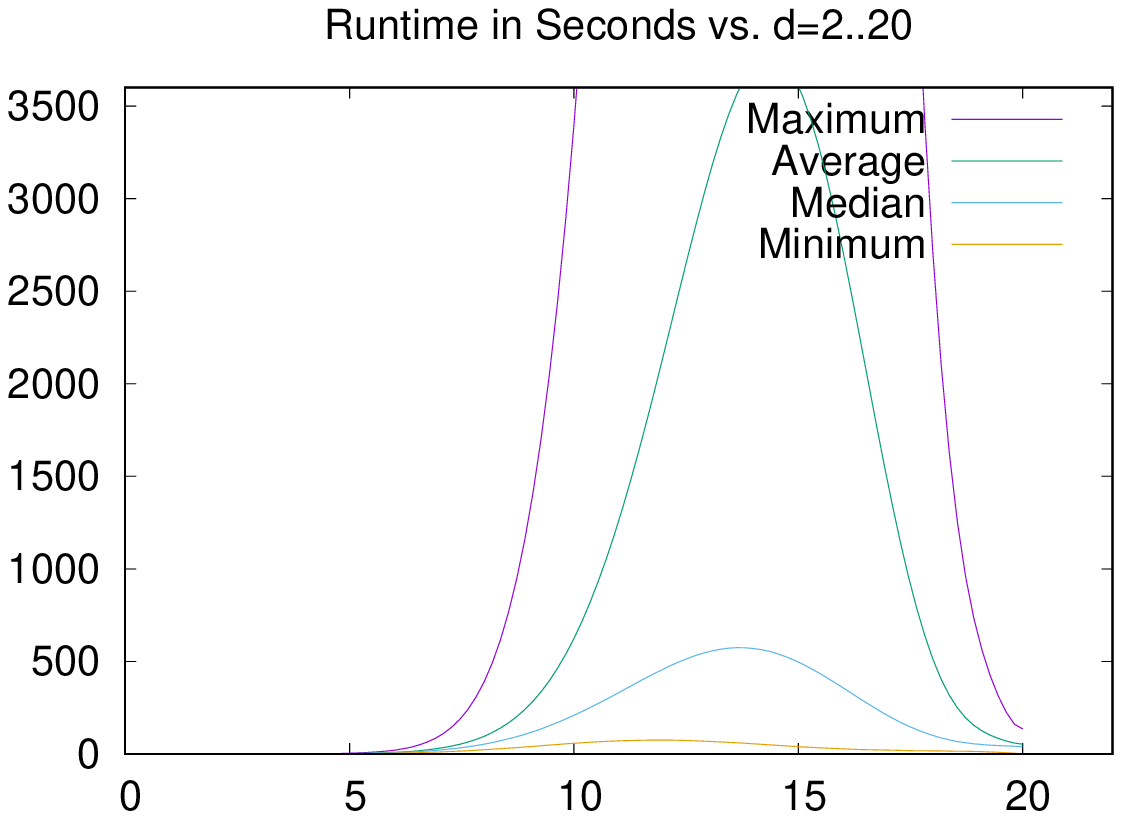}
\end{tabular}
\end{center}
\caption{Runtime scaling: checking \emph{satisfiability} vs computing \emph{intersection of solutions}\label{fig:sat-vs-cautious}}
\end{figure}

We generated $\QCN$ instances using model
${A(n=100,2\leq{d}\leq{20},s=6.5)}$~\cite{DBLP:journals/constraints/Nebel97},
where $n$ denotes the number of variables, $d$ the average degree, and
$s$ the average size (number of base relations) of a constraint of a
given instance.  For each $d \in \{2,\ldots,20\}$, we report runtimes
based on $10$ random instances because the runtime distribution is
\emph{heavy tailed}, i.e., the severity of outliers encountered
increases along the number of instances generated. As a consequence,
the maximum and average runtimes tend to infinity as can be seen from
the plots in Figure~\ref{fig:sat-vs-cautious}.  The graphs have been
smoothened using \system{gnuplot}'s option \emph{bezier}.

\begin{table}[t!]
\renewcommand{\sp}{\hspace{0.5em}}
\renewcommand{\b}[1]{\textbf{#1}}
\begin{center}
\begin{tabular}{|r|r@{\sp}r@{\sp}r@{\sp}r@{\sp}r@{\sp}r@{\sp}r@{\sp}r@{\sp}r@{\sp}r@{\sp}r|}
\hline
$d$ & 9 & 10 & 11 & 12 & 13 & 14 & 15 & 16 & 17 & 18 & 19 \\ \hline
Satisfiability
&   \b{4.7}
&  \b{34.9}
&  \b{60.9}
& \b{163.0}
& \b{180.7}
& \b{543.8}
& \b{157.3}
&  \b{38.0}
&  \b{32.5}
&   86.5
&   56.4 \\
Backbone
&     24.7
&     67.8
&    210.0
&    483.5
&    658.8
&   1488.4
&    223.0
&    382.9
&     64.6
&  \b{44.1}
&  \b{55.2} \\ \hline
\end{tabular}
\end{center}
\caption{Median runtimes for IA instances with $100$ variables \label{table:sat-vs-cautious}}
\end{table}

The graph on the left shows the runtime scaling for checking the
existence of a solution, and the graph on the right concerns the
computation of the intersection of solutions, which amounts to the
identification of \emph{backbones} for $\QCN$s~\cite{SJ19:ki-subm}. The
\system{clingo-dl} solver supports the computation of the intersection as
one of its command-line options. It is also worth noting a phase
transition around the value $d=14$ where instances turn from
satisfiable to unsatisfiable, which affects the complexity of
reasoning. Moreover, due to outliers, it is perhaps more informative
to check the median runtimes as given in Table
\ref{table:sat-vs-cautious}. 
It is clear that intersection of solutions computation is more
demanding, but the difference is not tremendous. Moreover, to contrast
the performance of our encoding with respect to \cite{BFB16:iclp}, we
note that only $10\%$ of $190$ instances exceeded the timeout of $300$
seconds (this same timeout was used in that work). In addition, the
experiments of \cite{BFB16:iclp} covered instances from $20$ to $50$
variables only and the encodings were already performing poorly by the
time $50$ variables were considered.  On the other hand, our encoding
still underperforms with respect to native QSTR tools and, at least as
far as satisfiability checking is concerned, the state-of-the-art
qualitative reasoner \system{gqr}~\cite{Gantner_gqr} tackles each of
the $190$ instances in a few seconds on average. To the best of our
knowledge, there is no native QSTR tool for calculating intersection
of solutions and in this way the advanced reasoning modes of the
\system{clingo-dl} solver enable new kinds of inference and for free,
since the same encoding can be used and no further implementation work
is incurred.

\begin{figure}[t]
\begin{center}
\begin{tabular}{@{}c@{}c@{}}
\includegraphics[width=0.50\textwidth]{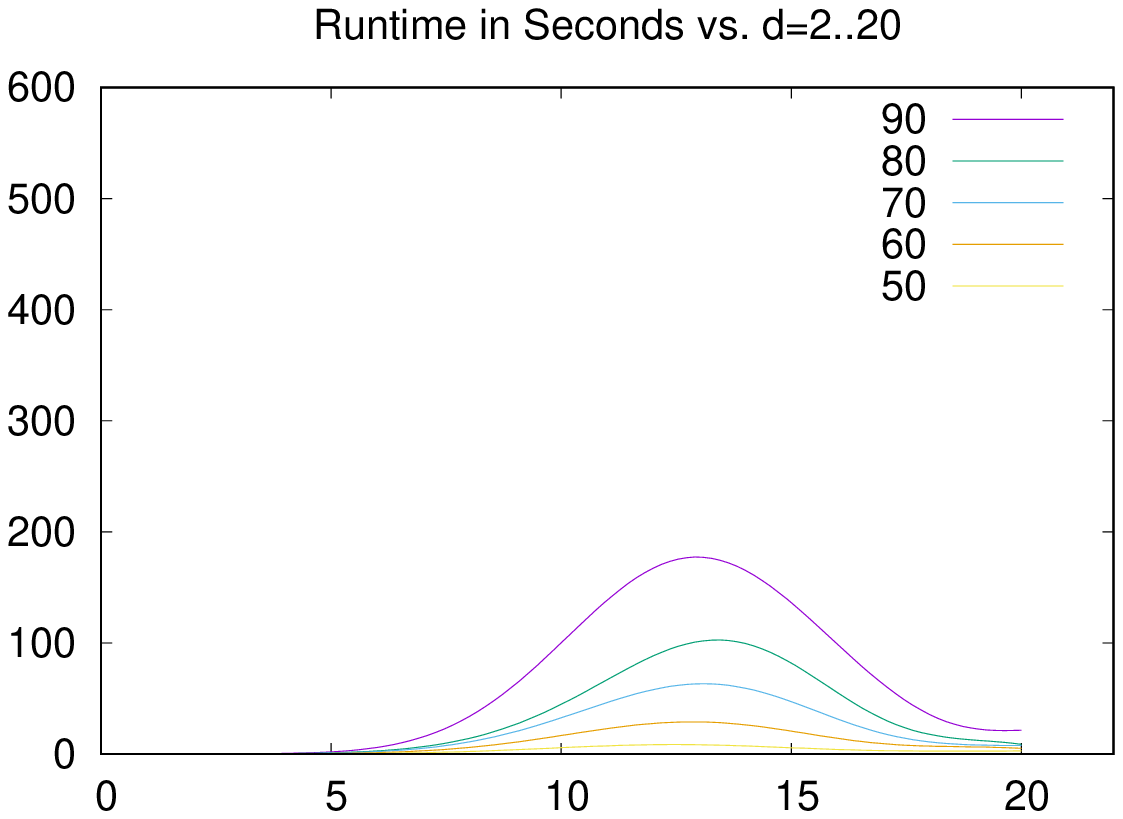}
&
\includegraphics[width=0.50\textwidth]{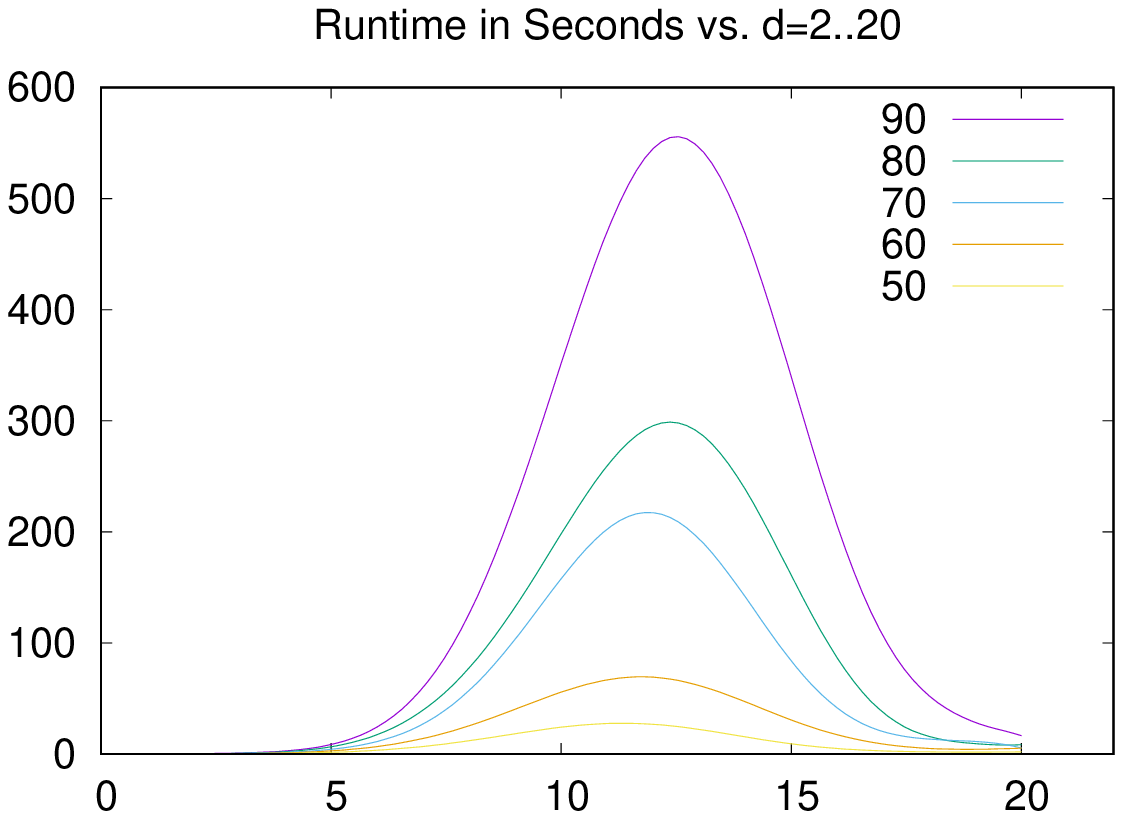}
\end{tabular}
\end{center}
\caption{Runtime scaling (median): computing intersection of solutions vs
computing union of solutions \label{fig:scale-easy}}
\end{figure}

Our second experiment studies the scalability of our ASP(DL) encoding
when the number of variables is gradually increased from $50$ to $90$.  The
results are illustrated in Figure \ref{fig:scale-easy}.  The plots on
the left illustrate the scaling of the backbone computation, i.e., the
intersection of solutions. It turned out that this kind of reasoning
is easier than computing the union of solutions, also known as the
\emph{minimum labeling problem}~\cite{Montanari74},
as depicted by the graphs on the right.
The random instances used so far are relatively easy, and for that
reason we take into consideration a modified scheme
${H(n,2\leq{d}\leq{20})}$~\cite{DBLP:journals/constraints/Nebel97}
that yields much harder network instances. The difference with respect
to model $A$ used above is that constraints are picked from a set of
relations expressible in 3-$\mathsf{CNF}$ when transformed into
first-order formulae. As a consequence, we are only able to analyze
instances up to $n=50$ variables in reasonable time.
Table~\ref{table:cautious-vs-brave}
shows the performance difference when computing the intersection and
the union of solutions. In most cases, the intersection of solutions
can be computed faster. Although $d=15$ is kind of an exception, its
significance is diminished by the most demanding instances
encountered: $8\,477$ vs $24\,199$ seconds spent on computing the
intersection and the union, respectively.

\begin{table}
\renewcommand{\sp}{\hspace{0.5em}}
\renewcommand{\b}[1]{\textbf{#1}}
\begin{center}
\begin{tabular}{|r|r@{\sp}r@{\sp}r@{\sp}r@{\sp}r@{\sp}r@{\sp}r@{\sp}r@{\sp}r@{\sp}r@{\sp}r|}
\hline
$d$ & 9 & 10 & 11 & 12 & 13 & 14 & 15 & 16 & 17 & 18 & 19 \\ \hline
Intersection
&    \b{4.8}
&    \b{8.7}
&   \b{19.8}
&   \b{50.8}
&  \b{122.3}
&  \b{940.7}
&    1738.0
&  \b{758.5}
&  \b{384.4}
&     258.0
&  \b{155.9} \\
Union
&      25.6
&      46.9
&     105.5
&     298.5
&    7226.3
&    5636.5
&  \b{749.8}
&    1585.5
&     438.9
&   \b{93.8}
&     169.3 \\ \hline
\end{tabular}
\end{center}
\caption{Median runtimes for IA instances with $50$ variables \label{table:cautious-vs-brave}}
\end{table}


\section{Conclusion and Future Work}
\label{sec:conclusion}

In this paper, we encoded qualitative constraint networks
($\QCN$s) based on Allen's Interval Algebra in ASP(DL), which is an
extension of answer set programming (ASP) by difference
constraints. Due to native implementation of such constraints as
propagators in the \system{clingo-dl} solver, the transitive effects of
relation composition are avoided when it comes to the space complexity
of representing $\QCN$ instances. This contrasts with existing
encodings in pure ASP~\cite{Li12:ictai,BFB16:iclp} and favors
computational performance, which rises to a new level due to our
ASP(DL) encoding. As regards other positive signs, it seems that the
presented encoding scales for other reasoning modes as well. Since
ASP encodings are highly elaboration tolerant, we expect that it
is relatively easy to modify and extend our basic encodings for
other reasoning tasks as well.
As regards future work, we aim to investigate more thoroughly the
performance characteristics of our ASP(DL) encoding, and to use it for
establishing collaborative frameworks among ASP-based and native QSTR
tools.


\end{document}